\documentclass{article}

\usepackage{microtype}
\usepackage{graphicx}
\usepackage{booktabs} 

\usepackage{hyperref}

\usepackage[accepted]{icml2020}

\usepackage{amsfonts}
\usepackage{bm}
\usepackage{bbm}
\usepackage{subfig}
\usepackage{amsmath}
\usepackage{amsthm}
\usepackage{algorithm}
\usepackage{algorithmic}
\newcommand{\mc}{\mathcal}
\newcommand{\bb}{\mathbb}
\newtheorem{thm}{Theorem}[section]
\newtheorem{prop}[thm]{Proposition}

\newcommand{\squishlist}{
 \begin{list}{$\bullet$}
  { \setlength{\itemsep}{0pt}
     \setlength{\parsep}{2pt}
     \setlength{\topsep}{2pt}
     \setlength{\partopsep}{0pt}
     \setlength{\leftmargin}{1em}
     \setlength{\labelwidth}{1em}
     \setlength{\labelsep}{0.5em} } }

     \newcommand{\squishend}{
  \end{list}  }
 
\icmltitlerunning{Value Variance Minimization for Learning Approximate Equilibrium in Aggregation Systems}

\begin{document}

\twocolumn[
\icmltitle{Value Variance Minimization for Learning Approximate Equilibrium in Aggregation Systems}




\begin{icmlauthorlist}
\icmlauthor{Tanvi Verma}{to}
\icmlauthor{Pradeep Varakantham}{to}
\end{icmlauthorlist}

\icmlaffiliation{to}{Singapore Management University, Singapore}
\icmlcorrespondingauthor{Tanvi Verma}{tanviverma.2015@phdcs.smu.edu.sg}

\icmlkeywords{Machine Learning}

\vskip 0.3in
]



\printAffiliationsAndNotice{}  

\begin{abstract} 

For effective matching of resources (e.g., taxis, food, bikes, shopping items) to customer demand, aggregation systems have been extremely successful. In aggregation systems, a central entity (e.g., Uber, Food Panda, Ofo) aggregates supply (e.g., drivers, delivery personnel) and matches demand to supply on a continuous basis (sequential decisions). Due to the objective of the central entity to maximize its profits, individual suppliers get sacrificed thereby creating incentive for individuals to leave the system. In this paper, we consider the problem of learning approximate equilibrium solutions (win-win solutions) in aggregation systems, so that individuals have an incentive to remain in the aggregation system. 

Unfortunately, such systems have thousands of agents and have to consider demand uncertainty and the underlying problem is a (Partially Observable) Stochastic Game.  Given the significant complexity of learning or planning in a stochastic game, we make three key contributions: (a) To exploit infinitesimally small contribution of each agent and anonymity (reward and transitions between agents are dependent on agent counts) in interactions, we represent this as a Multi-Agent Reinforcement Learning  (MARL) problem that builds on insights from non-atomic congestion games model; (b) We provide a novel variance reduction mechanism for moving joint solution towards Nash Equilibrium that exploits the infinitesimally small contribution of each agent; and finally (c) We provide detailed results on three different domains to demonstrate the utility of our approach in comparison to state-of-the-art methods. 
\end{abstract}

\section{Introduction}
Due to having more information about state (e.g., location) of supply (e.g., taxi/car drivers, delivery personnel) and current demand, aggregation systems provide a significant improvement in performance ~\cite{alonso2017demand,lowalekar2018online, verma2019correlated,bertsimas2019online} over decentralized decision making methods (e.g., individual taxi drivers using their memory and insights to find right locations when there is no customer on board) in serving uncertain customer demand. However, in aggregation systems, some suppliers can receive lower profits (e.g., due to servicing low demand and high cost areas) in maximizing overall profit for the centralized entity. This results in suppliers moving out and creating instability in the system. One way of addressing this instability is to learn equilibrium solutions for all the players (centralized entity and individual suppliers). 

Multi-Agent Reinforcement Learning (MARL)~\cite{littman1994markov} with an objective of computing equilibrium is an ideal model for representing aggregation systems. However, it is a challenging problem for multiple reasons: (i) typically there are thousands or tens of thousands of individual players; (ii) there is uncertainty associated with demand; and (iii) this is a sequential decision making problem, where decisions at one step have an impact on decisions to be taken at next step.

While there has been a significant amount of research on learning equilibrium policies in MARL problems~\cite{littman1994markov,watkins1992q,hu1998multiagent,littman2001friend}, most of them can only handle a few agents.  Recently, there have been Deep Learning based methods that can scale to large numbers of agents such as Neural Fictitious Self Play (NFSP)~\cite{heinrich2016deep} and Mean Field Q-Learning~\cite{pmlrv80yang18d}. However, neither of these approaches are able to exploit some of the key properties of aggregation systems (mentioned in the next paragraph) and as we show in the experimental results, perform worse than our approach. 

To address the computational complexity, we exploit three key aspects of aggregation systems. First, even though there are thousands of individual players, their contribution to overall social welfare is infinitesimal. Second, similar to congestion games, interactions among agents are anonymous (e.g.,  in traffic routing or network packet routing, the cost incurred by an agent is dependent on the number of other agents selecting the same path). Finally, as is typical in aggregation systems, centralized entities can provide guidance to the individual suppliers. 

Specifically, our key contributions are as follows: (a) We propose a Stochastic Non-atomic Congestion Games (SNCG) model to represent anonymity in interactions and infinitesimal contribution of individual agents for aggregation systems; (b) We then provide key theoretical properties of equilibrium in SNCG problems; (c) Most importantly, we then propose an MARL approach (based on insights in (b)) for SNCG problems that reduces variance in agent values to move joint solutions towards equilibrium solutions; and (d) We provide detailed experimental results on multiple benchmark domains from literature and compare against leading MARL approaches. 

\section{Motivating Problems}
Our work is motivated by MARL problems with large number of infinitesimally small agents , i.e., effect of single agent on the environment dynamics is negligible. Also, the interactions among the agents are anonymous. 

Car aggregation companies like  Uber, Lyft, Didi, Grab, Gojek etc. match car drivers to the customers demands. The individual drivers make sequential decisions to maximize their own long term revenue and they earn by competing for demand with each other.  Probability of a demand being assigned to a car is dependent on the number of other cars present in the origin location of the job and they can benefit by learning to move to advantageous locations.  Similarly, food delivery systems (Deliveroo, Ubereats, Foodpanda, DoorDarsh etc.) and grocery delivery systems (AmazonFresh, Deliv, RedMart etc) utilize services of delivery personnel to serve the food/groceries to the customers. 

Traffic routing is another example domain where travelers take sequential decisions to minimize their own overall travel time. Also, their travel time is affected by the congestion on the road network and a centralized traffic controller can provide guidance to drivers through information boards. 

\section{Related Work}
For contributions in this paper, the most relevant research is on computing equilibrium policies in MARL problems, which is represented as learning in stochastic games~\cite{shapley1953stochastic}. Minimax-Q \cite{littman1994markov} is one of the early equilibrium-based MARL algorithm that uses minimax rule to learn equilibrium policy in two-player zero-sum games. Nash-Q learning \cite{hu1998multiagent} is another popular algorithm that extends the classic single agent Q-learning \cite{watkins1992q} to general sum stochastic games. At each state, Nash-Q learning computes the Nash equilibria for the corresponding single stage game and uses this equilibrium strategy to update the Q-values. \cite{littman2001friend}  proposed Friend-or-Foe Q-learning (FFQ)  which has less strict convergence condition compared to Nash-Q. Another algorithm similar to Nash-Q learning is correlated-Q learning \cite{greenwald2003correlated} which uses value of correlated equilibria to update the Q-values instead of Nash equilibria. In fictitious self play (FSP) \cite{heinrich2015fictitious} agents learn best response through self play. FSP is a learning framework that implements fictitious play \cite{brown1951iterative} in a sample-based fashion. Unfortunately, all these algorithms are generally suited for a few agents and do not scale if number of agents is very large, which is the case in problems of interest in this paper. 

Recently, few deep learning based algorithms have been proposed to learn approximate Nash equilibrium. Neural fictitious self play (NFSP) \cite{heinrich2016deep} combines FSP with a neural network function approximation to provide a decentralized learning approach. Due to decentralization, NFSP is extremely scalable and can work on problems with many agents.  Mean field Q-learning (MFQ) \cite{pmlrv80yang18d} is a \textit{centralized learning decentralized execution} algorithm where individual agents learn Q-values of its interaction with average action of its neighbour agents. However, none of these approaches can directly exploit the key properties of aggregation systems (infinitesimal contribution of individual agents, anonymity in interactions, presence of a guiding centralized entity) to improve solution quality. As we demonstrate in our experimental results, our approach that benefits from exploiting these key properties of aggregation systems is able to outperform NFSP and MFQ with respect to quality of $\epsilon$-equilibrium solutions on multiple benchmark problem domains from literature.  

In this paper, we build on key results from non-atomic congestion games~\cite{roughgarden2002bad,roughgarden2007routing,fotakis2009structure,chau2003price,krichene2015online,bilancini2016strict} by accounting for transitional uncertainty.  While, there has been some research~\cite{angelidakis2013stochastic} on considering uncertainty in congestion games, the uncertainty considered there is in cost functions and not in state transitions. There has been other work~\cite{varakantham2012decision} that has considered congestion in the context of stochastic games. However, the focus there is on planning (and not learning) without a centralized entity and there is also an approximation on value function considered in that work. 

\section{Background: NCG}
In this section we provide a brief overview of Non-atomic Congestion Games (NCG) .

NCG has either been used to model selfish routing~\cite{roughgarden2002bad,roughgarden2007routing,fotakis2009structure} or resource sharing~\cite{chau2003price,krichene2015online,bilancini2016strict} problems. Though the underlying model is the same, there is a minor difference in the way the model is represented. Here we present a brief overview of NCG from the perspective of resource sharing problem as that is of relevance to contributions in this paper. For detailed exposition of NCG, we refer the readers to \cite{krichene2015online}.

In NCG, a finite set of resources $\mc{L}$ are shared by a set of players $\mc{X}$. To capture the infinitesimal contribution of each agent, the set $\mc{X}$ is endowed with a measure space: $(\mc{X}, \mc{M}, m)$. $\mc{M}$ is a $\sigma$-algebra of measurable subsets, $m$ is a finite Lebesgue measure and is interpreted as the mass of the agents. This measure is non-atomic, i.e., for an agent $x$, $m(\{x\}) = 0$. The set $\mc{X}$ is partitioned into $K$  populations, $\mc{X} = \mc{X}_1 \cup...\cup \mc{X}_{K}$. 

Each population type $k$ possesses a set of strategies $U_k$, and each strategy corresponds to a subset of the resources. Each agent selects a strategy, which leads to a joint strategy distribution, $\bm{a}$: 
$$\bm{a} = (f^u_k)_{u \in U_k, 1 \le k \le K} \textbf{ with } \sum_{u \in U_k} f^u_k = m(\mc{X}_k) \forall k$$ 
Here $f^u_k$ is the total mass of the agents from population $k$ who choose strategy $u$. The total consumption of a resource $l \in \mc{L}$ in a strategy distribution $\bm{a}$ is given by:
$$\phi^l(\bm{a}) = \sum_{k=1}^{K} \sum_{u \in U_k: l \in u} f^u_k$$ 
The cost of using a resource $l \in \mc{L}$ for strategy $\bm{a}$ is: $$c_l(\phi^l(\bm{a}))$$ where the function $c_l(.)$ represents cost of congestion and is assumed to be a non-decreasing continuous function. The cost experienced by an agent of type $k$ which selects strategy $u \in U_k$ is given by: $${\mc C}^u_k(\bm{a}) = \sum_{l \in u} c_l(\phi^l(\bm{a}))$$ 

A strategy $\bm{a}$ is Nash equilibrium if:
$$\forall k, \forall u, u' \in U_k: \textbf{ if } f^u_k > 0, \textbf{ then } {\mc C}^{u'}_{k}(\bm{a}) \ge {\mc C}^u_k(\bm{a})$$ Intuitively, it implies that the cost for any other strategy, $u'$ will be greater than or equal to the cost of strategy, $u$. In other words, it 
also implies that for a population $\mc{X}_k$, all the strategies with non-zero mass will have equal costs.

\section{Stochastic Non-atomic Congestion Games}
We propose Stochastic Non-atomic Congestion Game (SNCG) model to represent anonymity in interactions and infinitesimal agents in aggregation systems by extending non-atomic congestion games. Formally, SNCG is represented using the tuple: 

$\hspace{1in} \big< \mc{X, S, K}, \mc{A}, \mc{T}, \mc{R} \big>$
\squishlist
\item[$\mc{X}$:] Similar to NCG, $\mc{X}$ is the set of agents endowed with a measure space, $(\mc{X}, \mc{M}, m)$, where $\mc{M}$ is a $\sigma$-algebra of measurable subsets and $m$ is a finite Lebesgue measure. For an agent $x, \{x\}$ is a null-set and $m(\{x\})$ is zero. 

\item[$\mc{K}$:] is the set of local states of individual agents (e.g., location of a taxi).

\item[$\mc{S}$:] is the set of global states (e.g., distribution of taxis in the city).  The set of agents present in local state $k$ in global state $s$ is given by $\mc{X}^s_k$ and the mass of agents present in the local state, $k$ is given by $m(\mc{X}^s_k)$. The distribution of mass of agents is considered as the global state, i.e., $$s=<m(\mc{X}^s_1), m(\mc{X}^s_2),...,m(\mc{X}^s_{|\mc{K}|})>, \text{ with }$$ $$ \sum_{k=1}^{|\mc{K}|} m(\mc{X}^s_k) = 1 \forall s \in \mc{S}$$
The total mass of agents in any global state $s$ is 1. 
\item[$\mc{A}$:]  is the set of actions where $\mc{A}_k$ represents  the set of actions (e.g.,locations to move to) available to individual agents in the local state $k$.  $$\mc{A}= \{\mc{A}_k \}_{k \in \mc{K}}$$
Let $a(x)$ provides the action selected by agent $x$. We define $f^{u}_{k}(s) $ as the total mass of agents in $\mc{X}^s_k$ selecting action $u$ in state $s$, i.e. $\sum_{u \in \mc{A}_k} f^{u}_{k}(s)  = m(\mc{X}_k^s)$. If the agents are playing deterministic policies, $f^{u}_{k}(s)$ is given by
\begin{align}
f^{u}_{k}(s) = \int_{x \in \mc{X}^s_k} \mathbbm{1}_{(a(x)=u)} dm(x) \label{act-mass}
\end{align}

\item[$\mc{R}$]: is the reward function\footnote{Researchers generally use the term "cost" in the context of NCG. To be consistent with the MARL literature we use the term "reward". However, reward and cost can be used interchangeably by observing that reward is negative of cost.}.  The total mass of agents selecting action $u$ for a joint action $\bm{a}$ in state $s$ is given by
$$
\phi^{u}(\bm{a}) =\sum_{k=1; u \in \mc{A}_k}^{|\mc{K}|} f^{u}_{k}(s)
$$
Similar to the cost functions in NCG, the reward function is assumed to be a non-decreasing continuous function. The immediate reward is dependent on the mass of the agents selecting the same action. Also, all the agents which select action $u$ in local state $k$ receive equal reward\footnote{In aggregation systems, expected reward is equal for all the agents who perform the same action in a local state, i.e. who select to move to the same zone. } which is given by
$$ \mc{R}_k(s, \phi^{u}(\bm{a})) $$
\item[$\mathcal{T}$]: is the transitional probability of global states given joint actions. The global transition from the perspective of an individual agent $x$ is given by:
$$
\mc{T}(s'| s, \bm{a}) = p_k(k' | s, u, \bm{a}_{-x}) \cdot \mc{T}(s'_{-x}| s, \bm{a}) 
$$
$p_k(k' | s, u, \bm{a}_{-x})$ is the probability of moving to local state $k'$ when an agent $x \in \mc{X}^s_k$ takes action $u$ and the induced joint action by all the agents is $\bm{a}$. $\bm{a}_{-x}$ is the joint action induced by all the agents except $x$ and $s'_{-x}$ is the global state without agent $x$. 
\squishend

The policy of agent $x$ is denoted by $\pi_x$. We observe  that given a joint state $s$, an agent will play different policies based on its local state $k$ as the available actions for local states are different. Hence, $\pi_x$ can be represented as
\begin{align*}
\pi_x = (\pi_{xk}(s))_{s \in \mc{S}, k \in \mc{K}}  \text{  such that} \sum_{u \in \mc{A}_k} \pi_{xk}(u|s) = 1 
\end{align*}
We define $\Pi_k$ as the set of policies available to an agent in local state $k$, hence, $\pi_{xk}(s) \in \Pi_k  \forall x \in \mc{X}_k, \forall s \in \mc{S}$. $\bm{\pi} = (\pi_x)_{(x \in \mc{X})}$ is the joint policy of all the agents. 


Let $\gamma$ be the discount factor and $\rho_{\bm{\pi}}$ denotes the state-action marginals of trajectory distribution induced by the joint policy $\bm{\pi}$. We use $\rho_{x\bm{\pi}}$ to denote the local state-action trajectory distribution of agent $x$ induced by the joint policy $\bm{\pi} = (\pi_x, \bm{\pi}_{-x})$, where $\bm{\pi}_{-x}$ is the joint policy of other agents.
The value of agent $x$ for being in local state $k$ given the global state is $s$ and other agents are following policy $\bm{\pi}_{-x}$ is given by
{\small
\begin{align}
 v_{xk}(s, \pi_{xk}, \bm{\pi}_{-x})  &=  \bb{E}_{((s, \bm{a}) \sim \rho_{\bm{\pi}}, (k', u) \sim \rho_{x\bm{\pi}})} \Big[\sum_{t=0}^{\infty} \gamma^t \mc{R}_{k'}(s, \phi^u(\bm{a})) \Big] \nonumber\\ 
&\hspace{-0.2in}= \mc{R}_k(s, \phi^{\pi_{xk}(s)}(\bm{a})) \nonumber + \gamma \sum_{s'_{-x} \in \mc{S}_{-x}} \mc{T}(s'_{-x} | s, \bm{a}) \nonumber\\
&\hspace{0.1in} \sum_{k' \in \mc{K}} p_k(k'| s, \pi_{xk}(s), \bm{a}_{-x}) v_{xk'}(s', \pi_{xk'}, \bm{\pi}_{-x}) \label{pure-val}
\end{align}

\textbf{\em The goal in an SNCG is to compute an equilibrium joint strategy, where no agent has an incentive with respect to their individual value to unilaterally deviate from their solution. }

Here, we provide key properties of value function and equilibrium solution in SNCG that will later be used for developing a learning method for SNCGs. 

\begin{prop}\label{agent-value}
Values of other agents do not change if agent $x$ alone changes its policy. For any agent $y$ in any local state $k$:
{\small $$ v_{yk} (s, \pi_{yk}, \bm{\pi}_{-y} ) = v_{yk} (s, \pi_{yk}, \bm{\pi}'_{-y} )$$}
where $ \bm{\pi}_{-y} = \big(\pi_x, (\pi_z)_{z \in {\mc{X} \setminus \{x, y\}}}\big) $ and $\bm{\pi}'_{-y} = \big(\pi'_x, (\pi_z)_{z \in {\mc{X} \setminus \{x, y\}}}\big)$
\begin{proof} 
Adapting Equation~\ref{pure-val} for agent $y$ in local state $k$, we have: 
\begin{align}
v_{yk} (s, \pi_{yk}, \bm{\pi}_{-y} ) &=  \mc{R}_k(s, \phi^{\pi_{yk}(s)}(\bm{a})) \nonumber + \gamma \sum_{s'_{-x} \in \mc{S}_{-x}} \mc{T}(s'_{-y} | s, \bm{a}) \nonumber\\
&\hspace{0.2in} \sum_{k' \in \mc{K}} p_k(k'| s, \pi_{yk}(s), \bm{a}_{-y}) v_{yk'}(s', \pi_{yk'}, \bm{\pi}_{-y}) \label{eqn:3}
\end{align}

When policy of agent $x$ is changed, the main factor that is impacted in the RHS of the above expression is $\textbf{a}$ and due to that, the reward and transition terms can be impacted.  $\textbf{a}$ is solely dependent on $f_{k}^u(s)$ values and $f_{k}^u(s)$ values are dependent on the mass of agents taking action $u$ in local state $k$ and global state $s$ (Equation~\ref{act-mass}):
\begin{align}
 f^{u}_{k}(s) &= \int_{z \in \mc{X}^s_k} \mathbbm{1}_{(a(z)=u)} dm(z) \nonumber
 \intertext{If policy change makes agent $x$ move out of local state $k$ then the new mass of agents selecting action $u$ in $k$ is:}
 \tilde{f}^{u}_{k}(s) &= \int_{z \in {\mc{X}^s_k \setminus \{x\}}} \mathbbm{1}_{(a(x)=u)} dm(x) \nonumber
 \intertext{Since $f$ is primarily mass of agents (which is a Lebesgue measure), using the \textit{countable additivity} property of Lebesgue measure \cite{bogachev2007measure,hartman2014theory}, we have:}
 &= \int_{z \in {\mc{X}^s_k}} \mathbbm{1}_{(a(z)=u)} dm(z) -   \int_{z \in \{x\}} \mathbbm{1}_{(a(z)=u)} dm(z)
 \intertext{Since integral at a point in continuous space is 0 and mass measure is non-atomic, so  we have $\{x\}$ is a null set and $m(\{x\})=0$}
 &= \int_{z \in {\mc{X}^s_k}} \mathbbm{1}_{(a(z)=u)} dm(z) 
\end{align}

%

Since $f_{k}^{u}(s) = \tilde{f}^{u}_{k}(s)$, action, $\bm{a}$ remains same. Hence neither reward nor transition values change. Thus, RHS of Equation~\ref{eqn:3} remains same when $\bm{\pi}_{-y}$ is changed to $\bm{\pi}'_{-y}$ in the LHS. 
\end{proof}
\end{prop}

\subsection{Nash Equilibrium in SNCG}

A joint policy $\bm{\pi}$ is a Nash equilibrium if for all $k$ and for all $x \in \mc{X}_k$, there is no incentive for anyone to deviate unilaterally, i.e.
\begin{align}
v_{xk}(&s, \pi_{xk}, \bm{\pi}_{-x}) \ge v_{xk}(s, \pi'_{xk}, \bm{\pi}_{-x})\nonumber \\   & \forall s \in \mc{S}, \forall x \in \mc{X}_k, \forall k \in \mc{K}, \forall \pi_{xk}(s), \pi'_{xk}(s) \in \Pi_k \label{equilibrium-local}
\end{align}

\begin{prop} \label{equal-value}
Values of agents present in a local state are equal at equilibrium, i.e.,
\begin{align}
v_{xk}(&s, \pi_{xk}, \bm{\pi}_{-x}) = v_{yk}(s, \pi_{yk}, \bm{\pi}_{-y}), \nonumber \\   & \forall s \in \mc{S}, \forall x,y \in \mc{X}_k, \forall k \in \mc{K}, \forall \pi_{xk}(s), \pi_{yk}(s) \in \Pi_k
\end{align}
\begin{proof}
In the proof of Proposition~\ref{agent-value}, we showed that adding or subtracting one agent from a local state does not change other agent's values, as contribution of one agent is infinitesimal. Thus,
\begin{align}
v_{xk}(s, \pi_{xk}, \bm{\pi}_{-x}) &= v_{xk}(s, \pi_{xk}, \bm{\pi}) \textbf{ and also } \nonumber\\
v_{xk}(s, \pi'_{xk}, \bm{\pi}_{-x}) & = v_{xk}(s, \pi'_{xk}, \bm{\pi}) 
\end{align}
This implies that the value is dependent only the policy of the individual agent given its state and joint policy. Hence if agent $x$ in local state $k$ gets a highest value of $v_{xk}(s, \pi_{xk}, \bm{\pi})$ over all policies, then any other  agent $y$ in the same local state $k$ should get the same value. Otherwise, agent $y$ can swap to the same policy (all agents have access to the same set of policies in each local state) being used by $x$. Thus,
$$v_{xk}(s, \pi_{xk}, \bm{\pi}) = v_{yk}(s, \pi_{yk}, \bm{\pi})$$
and from the arguments in proof of Proposition~\ref{agent-value}, we have

$\hspace{0.65in} v_{xk}(s, \pi_{xk}, \bm{\pi}_{-x}) = v_{yk}(s, \pi_{yk}, \bm{\pi}_{-y})$
\end{proof}
\end{prop}

When there are multiple types of agents, we can provide a similar proof that values of same type of agents would be equal in a local state at equilibrium.

While SNCG model is interesting, it is typically hard to get the complete model before hand. Hence, we pursue a multi-agent learning approach to compute high-quality and fair joint policies in SNCG problems. 

\section{Value Variance Minimization Q-learning, VMQ}

We now provide a learning based approach for solving SNCG problems by utilizing Proposition~\ref{equal-value} in a novel way. As argued in Proposition \ref{equal-value}, the values of all the agents\footnote{Values of all the agents of same type in a local state are equal if there are multiple types of agent population present in the system.} present in any local state are equal at equilibrium. However please note that the converse is not true, i.e., even if the values of agents in local states are equal, the policy is not guaranteed to be an equilibrium policy. For a joint policy to be an equilibrium policy, agents should also be playing their best responses in addition to having values of agents in same local states being equal. 

\textbf{\em This is an ideal insight for computing equilibrium solutions in aggregation systems, as the centralized entity can focus on ensuring values of agents in same local states are (close to) equal by minimizing variance in values, while the individual suppliers can focus on computing best responses.}

VMQ is a \textit{centralized training decentralized execution} algorithm which assumes that during training a centralized entity has the access to the current values of the agents. The role of the central entity is to ensure that the exploration of individual agents moves towards a joint policy where the variance in values of agents in a local state is minimum. The role of the individual agents is to learn their best responses to the historical behavior of the other agents based on guidance from central entity. 

Algorithm \ref{algo-VMQ} provides detailed steps of the learning:
\squishlist
\item{Central agent suggests joint action $\bm{a}^c$ based on the joint policy it has estimated to all the individual agents. Line 11 of the algorithm shows this step.}  For the central agent, we consider a policy gradient framework to learn the joint policy. $\sigma(s, \bm{a})$ is the long term mean variance in the values of agents in all the local states if they perform joint action $\bm{a}$. 

We define two parameterized functions: joint policy function $\mu(s; \theta_{\mu})$ and variance function $\sigma(s, \bm{a}; \theta_\sigma)$. Since the goal is to minimize variance, we will need to update joint policy parameters in the negative direction of the gradient of $\sigma(s, \bm{a})$. Hence, policy parameters $\theta_{\mu}$ can be updated in the proportion to the gradient $-\nabla_{\theta_{\mu}}\sigma(s, \mu(s;\theta_{\mu}); \theta_\sigma)$. Using chain rule, the gradient of the policy will thus be
\begin{align}
- \nabla_{\theta_{\mu}}\sigma(s, &\mu(s;\theta_{\mu}); \theta_\sigma) = \nonumber \\ &-\nabla_{\theta_{\mu}}\mu(s;\theta_{\mu})\nabla_{\bm{a}}\sigma(s, \bm{a}; \theta_\sigma)|_{\bm{a} = \mu(s;\theta_{\mu})} \label{gradient}
\end{align}

\item{Individual agents either follow the suggested action with $\epsilon_1$ probability or play their best response policy with $1-\epsilon_1$ probability. While playing the best response policy, the individual agents explore with $\epsilon_2$ probability (i.e. $\epsilon_2$ fraction of $(1-\epsilon_1)$ probability) and with the remaining probability (($1-\epsilon_2$) fraction of $(1-\epsilon_1)$) they play their best response action. Line 13 shows this step.} The individual agents $x$ maintain a network $Q_x(s, k, u; \theta_x)$ to approximate the best response to historical behavior of the other agents in local state $k$ when global state is $s$. 
\item{Environment moves to the next state. All the individual agents observe their individual reward and update their best response values. Central agent observes the true-joint action $\bm{a}$ performed by the individual agents. Based on the true joint-action and variance in the values of agents, the central agent updates its own learning. }
\squishend

As common with deep RL methods \cite{mnih2015human,foerster2017stabilising}, replay buffer is used to store experiences ($\mc{J}$ for the central agent and $\mc{J}_x$ for individual agent $x$) and target networks (parameterized with $\theta'$) are used to increase the stability of learning. We define $\mc{L}_{\theta_\sigma}$, $\mc{L}_{\theta_\mu}$ and $\mc{L}_{\theta_x}$ as the loss functions of $\sigma$, $\mu$ and $Q_x$ networks respectively. The loss values are computed based on mini batch of experiences as follows
	\begin{align}
	&\mc{L}_{\theta_{\sigma}} = \mathbb{E}_{(s,\nu,\bm{a},s') \sim \mathcal{J}}\Big[ \Big( \nu + \gamma \cdot \sigma(s', \mu(s'; \theta'_{\mu})) - \sigma( s, \bm{a}; \theta_\sigma)\Big)^2\Big]  \label{varloss} \\
&\mc{L}_{\theta_{\mu}} = \mathbb{E}_{(s) \sim \mathcal{J}}\Big[ -\nabla_{\theta_{\mu}} \mu(s;\theta_{\mu}) \nabla_{\bm{a}} \sigma(s, \mu(s; \theta'_{\mu});\theta_\sigma) \Big] \label{ploss} \\
&\mc{L}_{\theta_x} = \mathbb{E}_{(s,k,u, r, s', k') \sim \mc{J}_x}\Big[ \big(r + \gamma \cdot \max_{u'} Q_x(s', k', u'; \theta'_x) \nonumber \\ 
& \quad \quad \quad \quad \quad \quad \quad \quad \quad \quad \quad \quad - Q_x(s, k, u; \theta_x) \big)^2\Big] \label{qloss}
	\end{align}
$\mc{L}_{\theta_\sigma}$ and $\mc{L}_{\theta_x}$ are computed based on TD error	\cite{sutton1988learning} whereas $\mc{L}_{\theta_\mu}$ is computed based on the gradient provided in Equation \ref{gradient}.

\begin{algorithm}
\caption{VMQ}
\label{algo-VMQ}
\begin{algorithmic}[1]
	\STATE {Initialize replay buffer $\mc{J}$, action-variance network $\sigma(s, \bm{a};\theta_\sigma)$, policy network $\mu(s;\theta_{\mu})$ and corresponding target networks with parameters $\theta'_\sigma$ and $\theta'_{\mu}$ respectively for the central agent}
\STATE {Initialize replay buffer $\mc{J}_x$, action-value network $Q_x(s,k,u; \theta_x)$ and corresponding target network with parameter $\theta'_x$ for all the individual agents $x$}
\WHILE{not converged}
\FOR{ $k \in \mc{K}$}
	\FOR{$x \in \mc{X}_k$}
	\STATE{compute value of $x$, $v_{xk} = max_uQ_x(s,k,u;\theta_x)$}
	\ENDFOR
\STATE{Compute $\nu_k$, variance in $v_{xk}$ values for $x \in \mc{X}_k$}
\ENDFOR
\STATE{Compute mean variance $\nu = \dfrac{1}{|\mc{K}|} \sum_{k \in \mc{K}} \nu_k$}
\STATE{Compute suggested joint action by the central entity $\bm{a}^c \leftarrow \mu(s,\theta^{\mu})$.\label{suggested_act}}
	\FOR {all $k \in \mc{K}$ and for all agent $x \in \mc{X}_k$ }
	\STATE{with probability $\epsilon_1$, \\ 
	$\quad \quad u_x \leftarrow \text{sample from }\bm{a}^c $ \\
	with remaining probability $1-\epsilon_1$ \\
	$\quad \quad u_x \leftarrow  \epsilon_2\text{-greedy} (Q_x)  $\label{indi-exploration}}
		\STATE{Perform action $u_x$ and observe immediate reward $r_x$ and next local state $k'$}
	\ENDFOR
	\STATE{Compute true joint action $\bm{a}$ and observe next state $s'$}
	 \STATE{Store transition $(s, \nu, \bm{a}, s')$ in  $\mc{J}$ and respective transitions $(s, k_x, u_x, r_x, s', k'_x)$ in $\mc{J}_x$ for all agents $x$}
	\STATE{Periodically update the network parameters by minimizing the loss functions provided in Equations \ref{varloss}-\ref{qloss} }
\STATE{Periodically update the target network parameters}
\ENDWHILE
\end{algorithmic}
\end{algorithm} 
\begin{figure}
\centering
\includegraphics[scale=0.35]{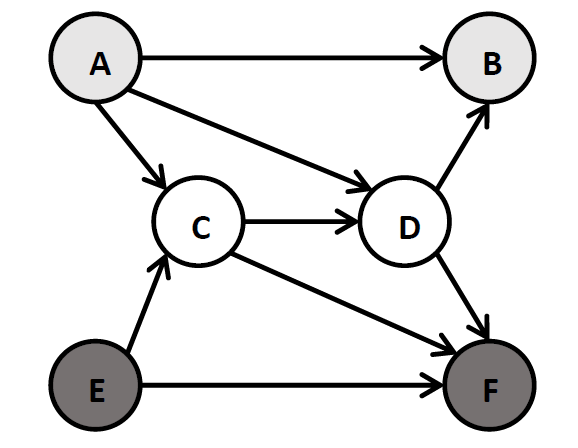}
\caption{Routing network}
\label{packet}
\end{figure}
\begin{figure}
\centering
\subfloat[Population $\mc{X}_1$ \label{p1}]{\includegraphics[scale=0.22]{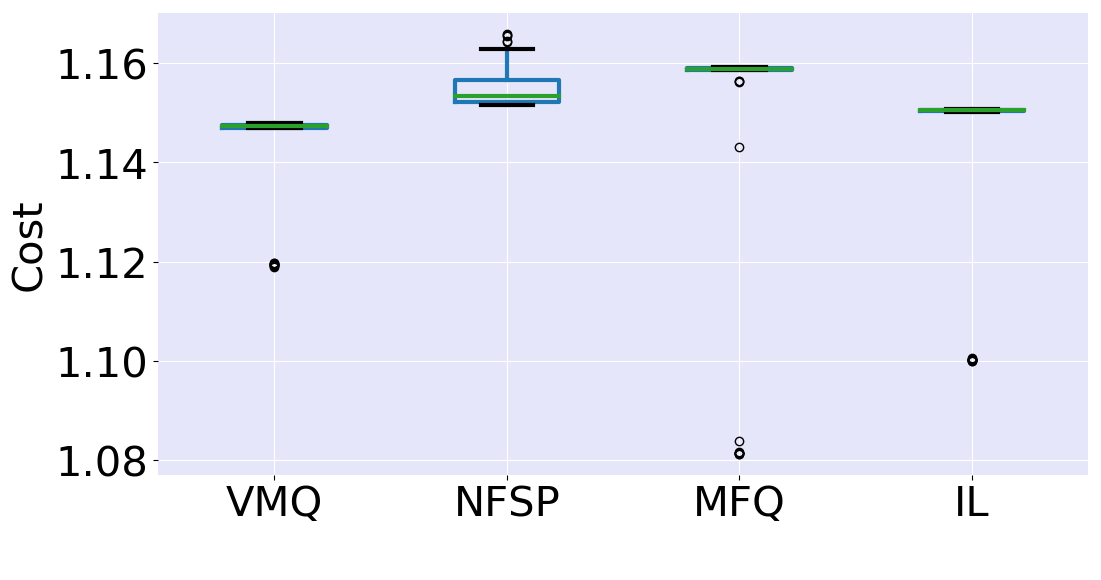}} \\
\subfloat[Population $\mc{X}_2$ \label{p2}]{\includegraphics[scale=0.22]{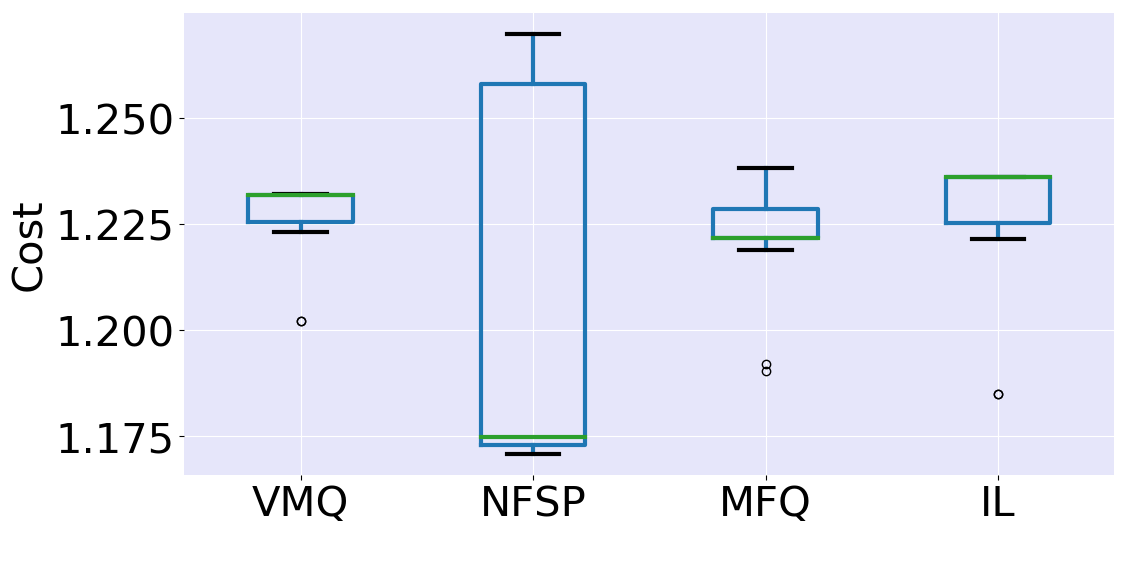}} 
\caption{Variance in costs of agents for packet routing example.}
\label{packet-routing-plot}
\end{figure}

\section{Experiments}

We perform experiments on three different domains, a single stage packet routing \cite{krichene2014learning}, muti-stage traffic routing \cite{wiering2000multi}, taxi simulator based on real-world and synthetic data set \cite{verma2019entropy,verma2019correlated}. In all these domains there is a central agent that assists (or provides guidance to) individual agents in achieving equilibrium policies. For example, a central traffic controller can provide suggestions to the individual travelers where as aggregation companies can act as a central entity for the taxi domain.

As argued in Proposition \ref{equal-value}, for SNCG the values of all the agents in a local state would be the same or variance in their values should be zero. Hence, we use variance in the values of all the agents as comparison measure (we use boxplots to show the variance). We compare with three baseline algorithms: Independent Learner (IL), neural fictitious self play (NFSP) \cite{heinrich2016deep} and mean-field Q-learning (MFQ) \cite{pmlrv80yang18d}.  IL is a traditional Q-Learning algorithm that does not consider the actions performed by the other agents. Similar to VMQ, MFQ is also a \textit{centralized training decentralized execution} algorithm and it uses joint action information at the time of training. However, NFSP is a self play learning algorithm and learns from individual agent's local observation. Hence, for fair comparison, we provide joint action information to NFSP as well. As mentioned by \cite{verma2019correlated}, we also observed that the original NFSP without joint action information performs worse that NFSP with joint action information. We use the best results for NFSP. 

Our neural network consisted of one hidden layer with 256 nodes. We also used dropout layer between hidden and output layer to prevent the network from overfitting. We used Adam optimizer for all the experimental domains. Learning rate was set to 1e-5 for all the experiments. For all the individual agents, we performed $\epsilon$-greedy exploration and it was decayed exponentially. Training was stopped once $\epsilon$ decays to 0.05. In all the experiments, each individual agent maintained a separate neural network. We experimented with different values of aniticipatory parameter for NFSP, we used 0.1 for Taxi Simulator and 0.8 for the remaining two domains which provided the best results.

\subsection{Packet Routing}

We first performed experiments with a single stage packet routing game \cite{krichene2014learning}. Two population of agents $\mc{X}_1$ and $\mc{X}_2$ of mass 0.5 each share the network given in Figure \ref{packet}. The first population sends packets from node $A$ to node $B$, and the second population sends from node $E$ to node $F$. Paths $AB, ACDB, ADB$ are available to agents in $\mc{X}_1$ whereas paths $EF, ECDF, ECF$ are available to agents in $\mc{X}_2$. The cost incurred on a path is sum of costs on all the edges in the path. The costs functions for the edges when mass of population on the edge is $\phi$ are given by:
$c_{AB}(\phi) = \phi + 2 $, $c_{AC}(\phi) = \phi /\ 2,  c_{AD}(\phi) = \phi, 
c_{DB}(\phi) = \phi /\ 3, c_{CD}(\phi) = 3\phi, c_{EC}(\phi) = 1 /\ 2,
c_{CF}(\phi) = \phi , c_{DF}(\phi) = \phi /\ 4, c_{EF}(\phi) = \phi + 1$
\small{
\begin{center}
\begin{table}
\caption{Comparison of policies and $\epsilon$ values for packet routing example}
\begin{tabular}{ | c | c | c |}
\hline 
Method & policy & $\epsilon$ value \\ \hline
Equilibrium Policy &((0, 0.187, 0.813),& 0\\ 
&(0.223, 0.053, 0.724))& \\ \hline
VMQ & ((0, 0.180, 0.820), & 0.07 \\ 
&(0.220, 0.040, 0.740))& \\ \hline
NFSP & ((0.004, 0.116, 0.88), &  0.792\\ 
&(0.01, 0.164, 0.826))& \\ \hline
MFQ & ((0, 0.162, 0.838),  &0.15\\ 
&(0.220, 0.040, 0.740))&  \\ \hline
IL & ((0.055, 0.176, 0.769),  &0.971\\ 
&(0.217, 0.088, 0.695))&  \\ \hline
\end{tabular}
\label{eq-policy}
\end{table}
\end{center}}
\begin{figure}
\centering
\subfloat[Population $\mc{X}_1$ \label{msp1}]{\includegraphics[scale=0.22]{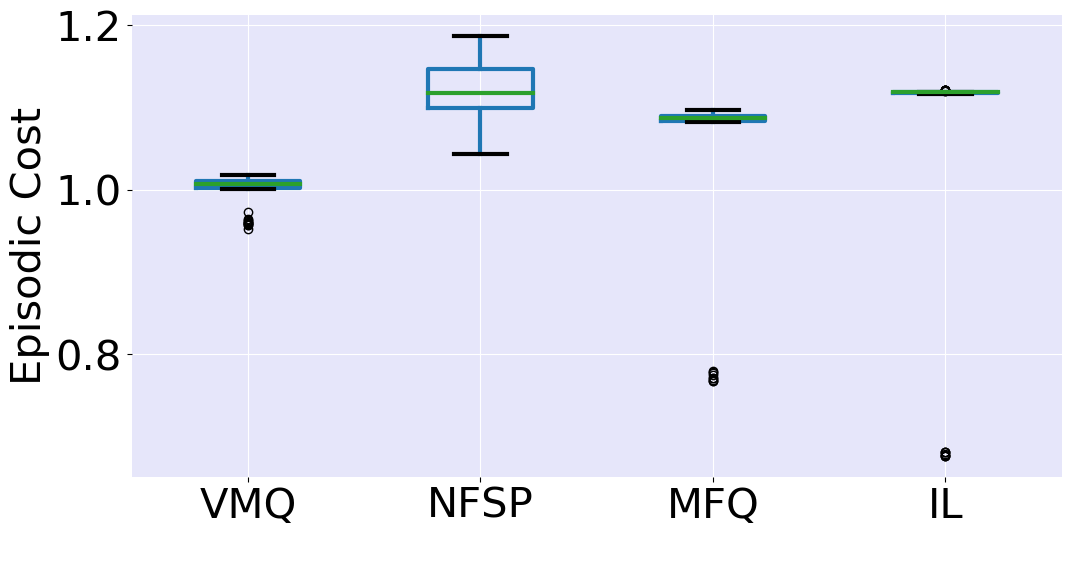}} \\
\subfloat[Population $\mc{X}_2$ \label{msp2}]{\includegraphics[scale=0.22]{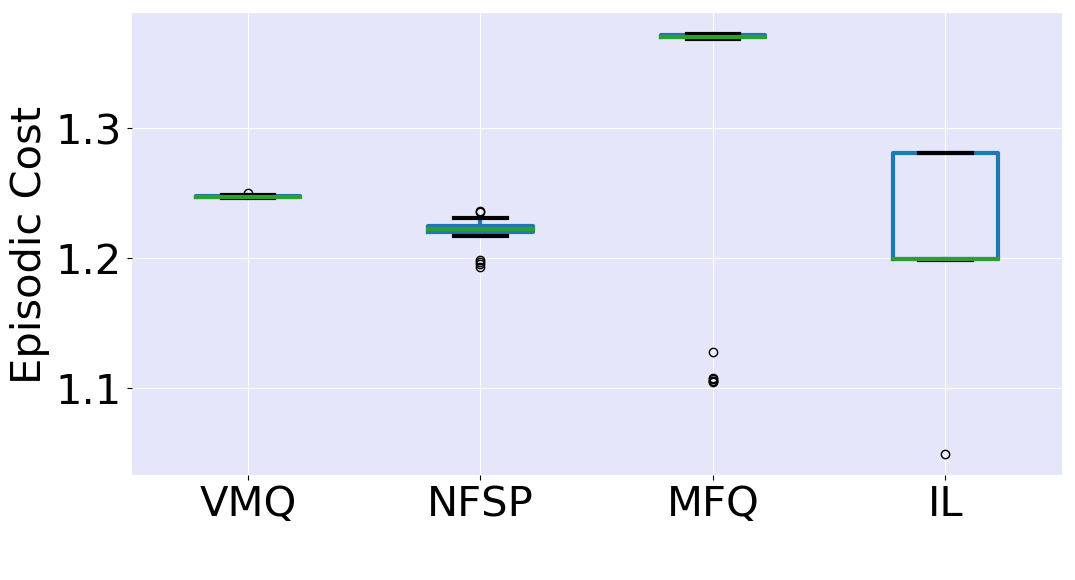}} 
\caption{Variance in values of agents for multi-stage traffic routing}
\label{ms-routing}
\end{figure}
\begin{figure}
\centering
\subfloat[Mean reward of agents \label{agent-mean}]{\includegraphics[scale=0.22]{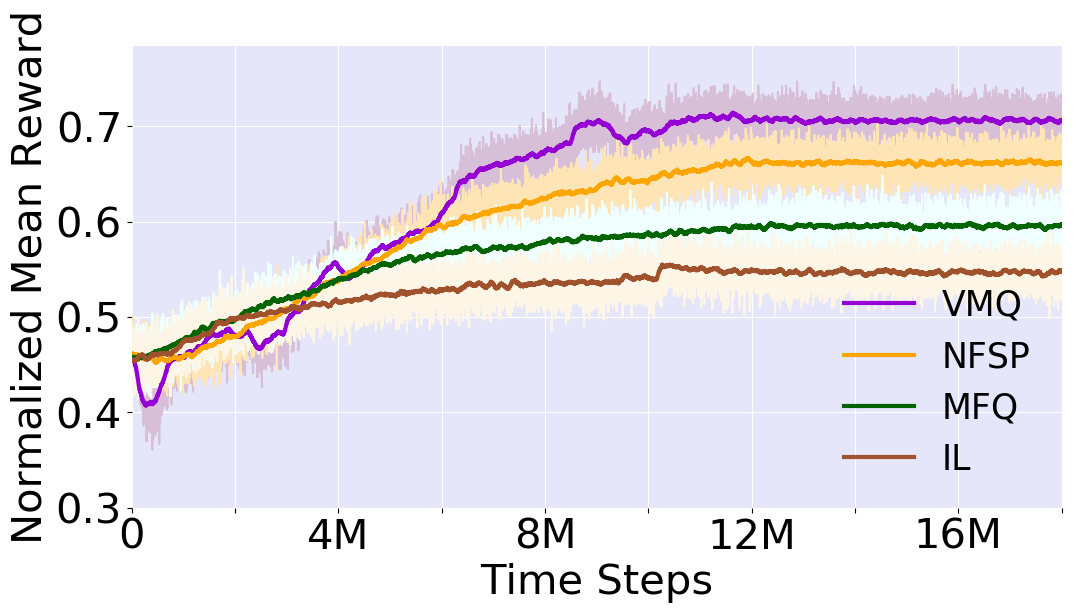}} \\
\subfloat[Variance in the values of agents \label{agent-var}]{\includegraphics[scale=0.22]{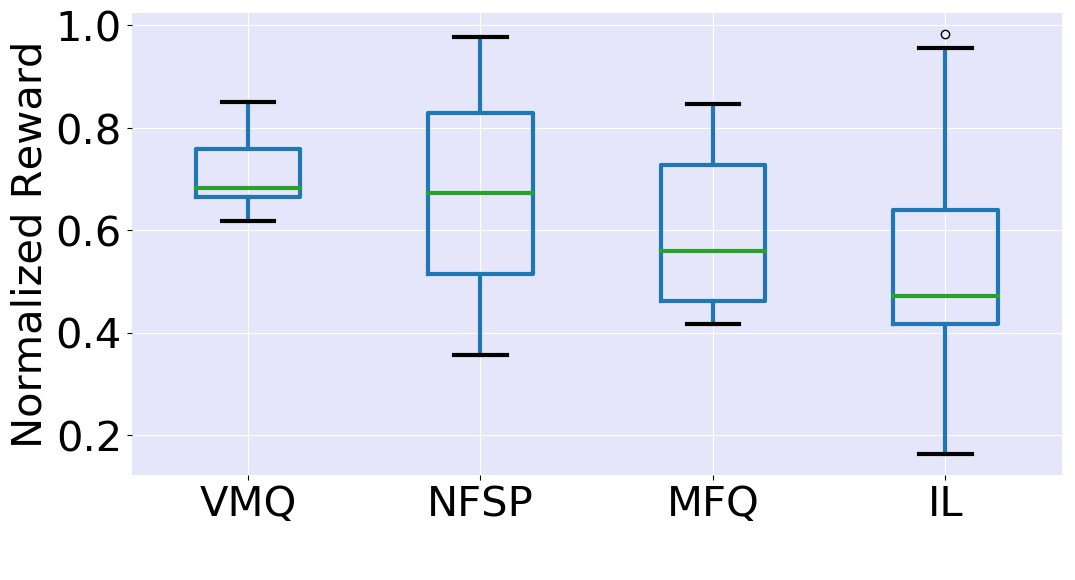}}
\caption{Taxi simulator using real-world data set}
\label{real-world}
\end{figure}
\begin{figure*}
    \centering
    \subfloat[DAR=0.4 \label{dar4mean}]
    {\includegraphics[scale=0.22]{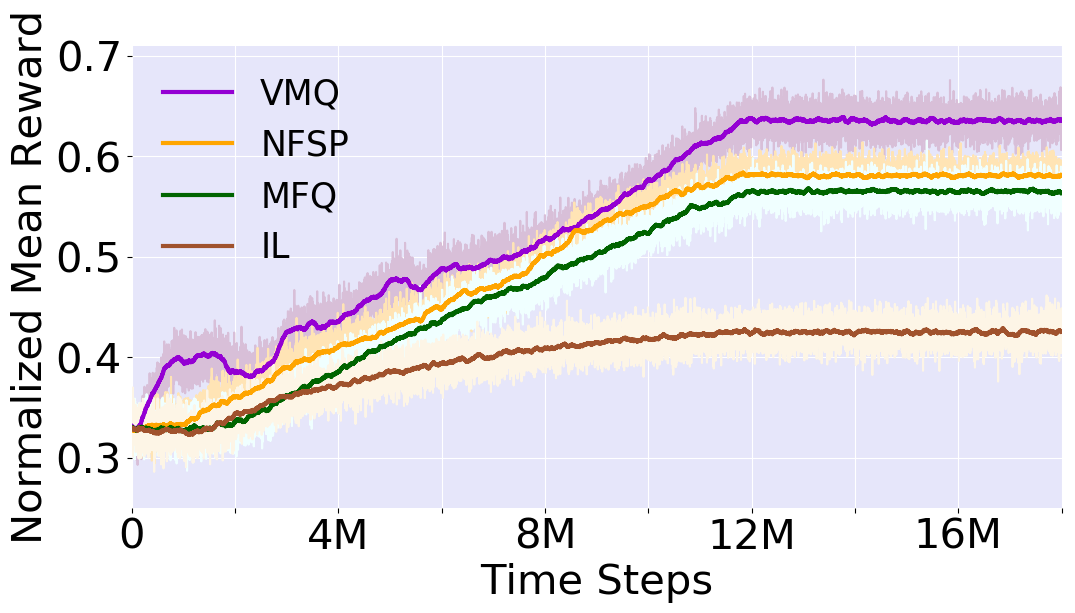}} 
    \subfloat[DAR=0.5\label{dar5mean}]
    {\includegraphics[scale=0.22]{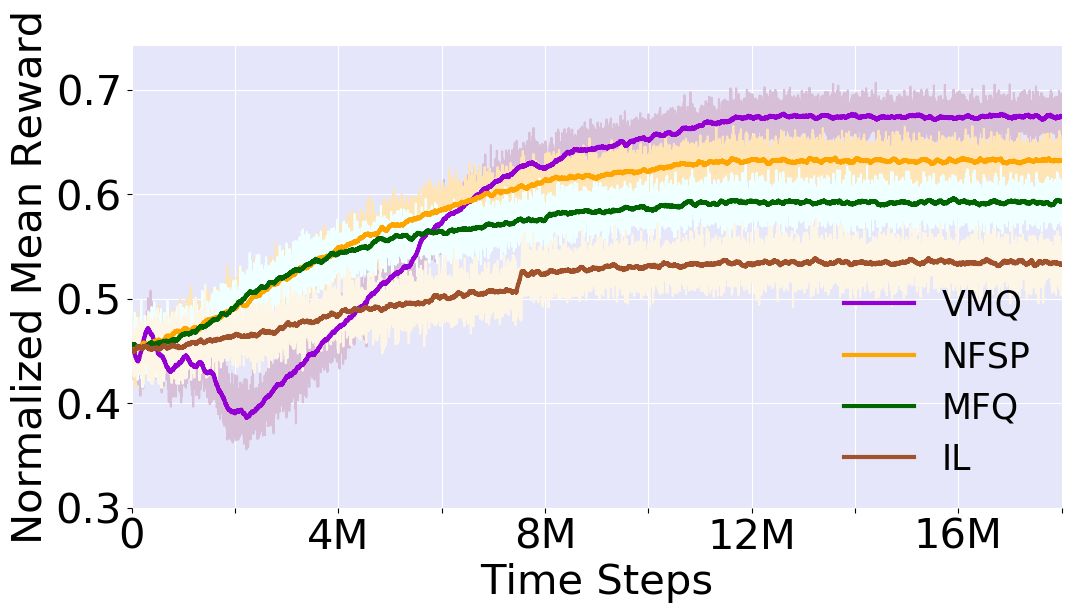}}
    \subfloat[DAR=0.6\label{dar6mean}]
    {\includegraphics[scale=0.22]{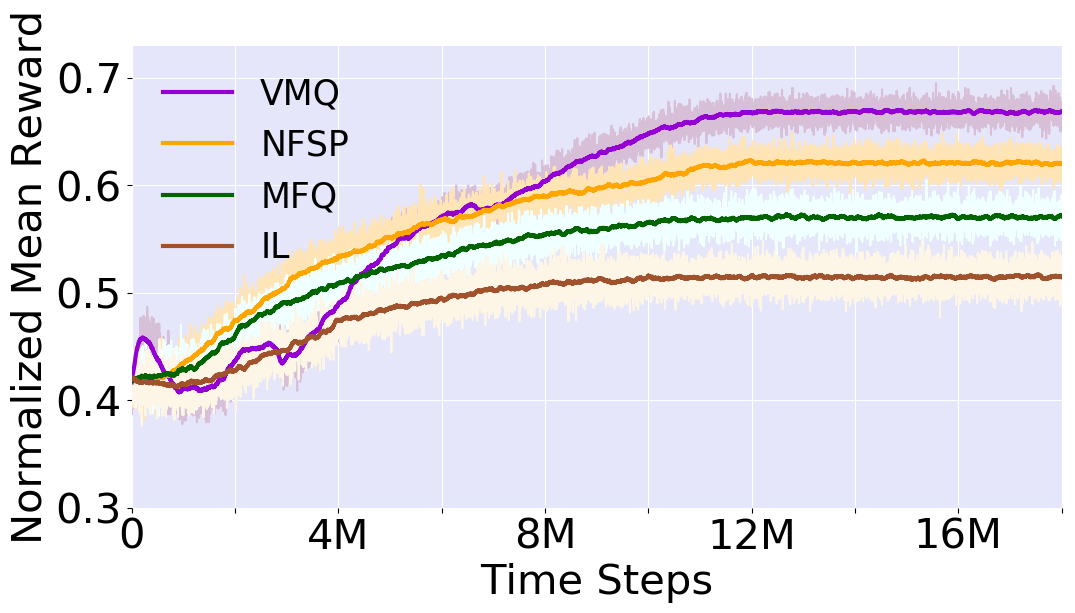}}\\
     \subfloat[DAR=0.4\label{dar4var}]
 {\includegraphics[scale=0.22]{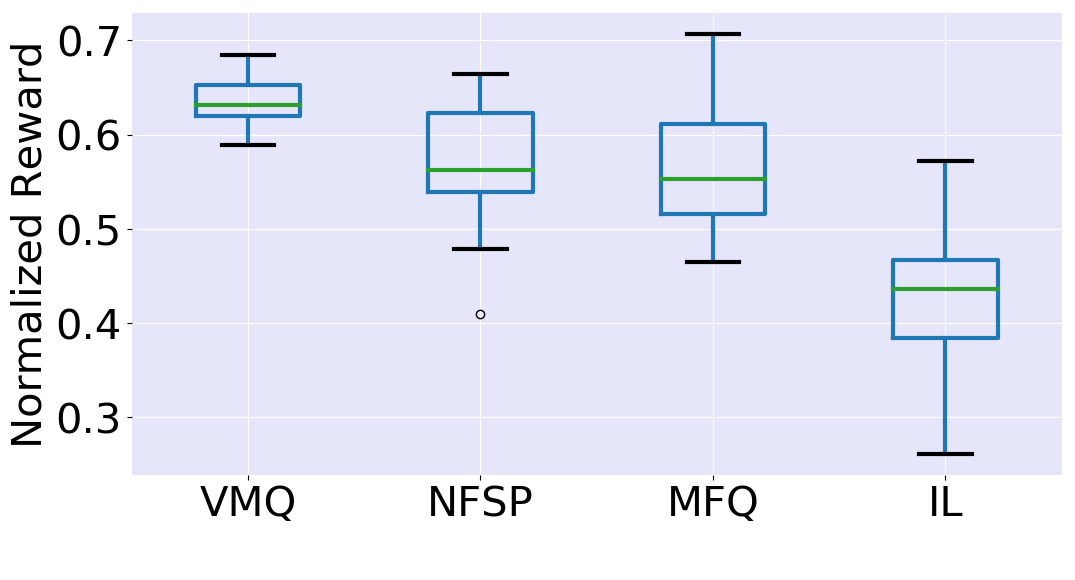}} 
 \subfloat[DAR=0.5\label{dar5var}]
 {\includegraphics[scale=0.22]{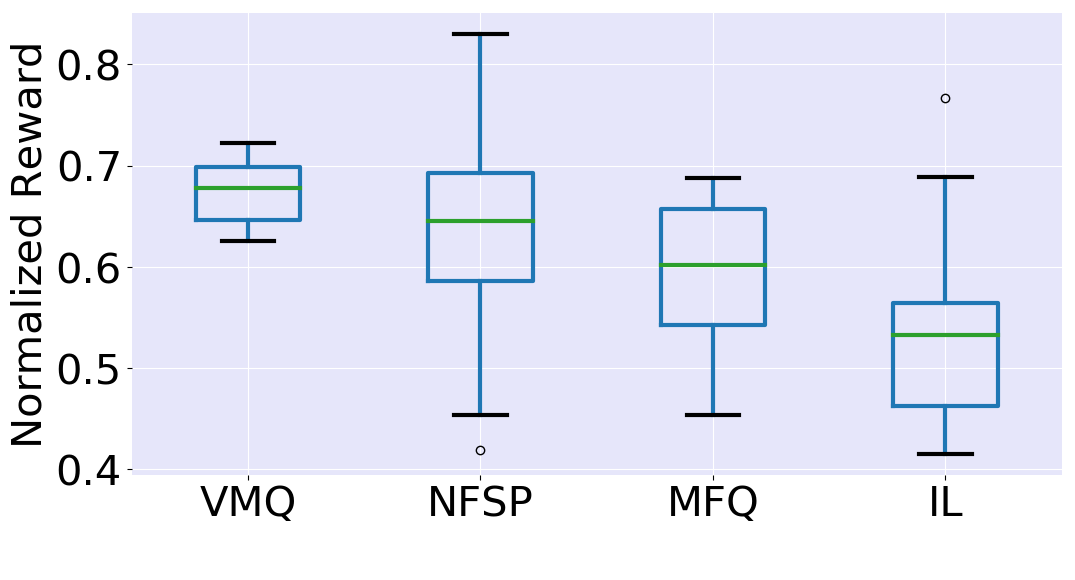}} 
 \subfloat[DAR=0.6\label{dar6var}]
 {\includegraphics[scale=0.22]{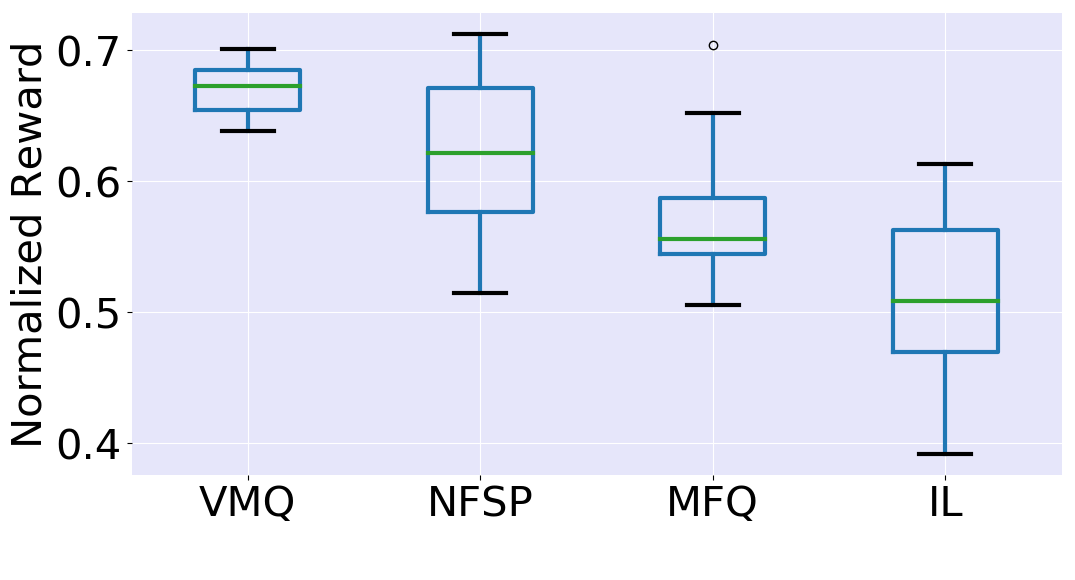}} 
    \caption{Mean reward of agents and variance in individual revenues comparison for taxi domain built from synthetic data set}  
    \label{synthetic}
\end{figure*}

If the cost functions are known, equilibrium policy can be computed by minimizing Rosenthal potential function \cite{rosenthal1973class}. We use equilibrium policy and costs on paths computed by minimizing potential function to compare quality of the equilibrium policy learned. We performed experiments with 100 agents of each type. We also compute $\epsilon$ values of the learned policy, which is the maximum reduction in the cost of an agent when it changes its policy unilaterally. Table \ref{eq-policy} compares the policies and $\epsilon$ values where the first row contain values computed using potential minimization method. The policy is represent as $
((\pi^{AB}_1, \pi^{ACDB}_1,\pi^{ADB}_1), (\pi^{EF}_2, f^{ECDF}_2,\pi^{ECF}_2))$,
where $\pi^p_i$ is the fraction of mass of population of type $i$ selecting path $p$.  We see that the VMQ policy is closest to the equilibrium policy and $\epsilon$ value is also lowest as compared to NFSP, MFQ and IL.

The equilibrium cost on paths as computed by the potential minimization method are: $
AB = 2, ACDB = ADB = 1.14,
EF = ECDF = ECF = 1.22$, i.e. at equilibrium agents in population $\mc{X}_1$ incur a cost of 1.14 whereas cost for agents in population $\mc{X}_2$ is 1.22.
Figures \ref{p1} and \ref{p2} provide variance in costs of agents for population $\mc{X}_1$ and $\mc{X}_2$ respectively. We can see that not only variance in the costs of agents is minimum for VMQ but the values are also very close to the equilibrium values computed using potential function minimization method. 

\subsection{Multi-Stage Traffic Routing}
We use the same network provided in Figure \ref{packet} to depict a traffic network where two population of agents $\mc{X}_1$ and $\mc{X}_2$ navigate from node $A$ to node $B$ and from node $E$ to node $F$ respectively. Unlike to the packet routing example, agents decide about their next edge at every node. Available edges to population type at every node remains the same as explained in the previous example. As the decision is made at every node, the  domain is an example of SNCG where agents make a sequence of decision to minimize their long term cost. Hence, the values of agents from a population at a given node would be equal at equilibrium.

In this example, agents perform episodic learning and the episode ends when the agent reach their respective destination nodes. The distribution of mass of population over all the nodes is considered as state. We perform experiments with 100 agents of each type. Figures \ref{msp1} and \ref{msp2} show the variance in values of both the population. Similar to the packet routing domain, the variance is minimum for VMQ. Furthermore, we notice that for both single-stage and multi-stage cases, the values of agents from $\mc{X}_2$ is affected only by their own aggregated policy and fraction of agents from $\mc{X}_1$ selecting path $AC$. However, for agents from $\mc{X}_1$, the values would be different from single-stage case. For example, agents selecting path $ACDB$ and $ADB$ would reach the destination node at different time steps and hence cost of agents on edge $DB$ would be different from the single-stage case. Hence we can safely assume that the equilibrium value of agents from $\mc{X}_2$ would be 1.22 as computed for the single-stage case which is the value for VMQ as shown in Figure \ref{msp2}. 

\subsection{Taxi Simulator}
Inspired from \cite{verma2019entropy} we build a taxi simulator based on both real-world and synthetic data set. Using GPS data of a taxi-fleet, the map of the city was divided into multiple zones (each zone is considered as a local state) and demand between any two zones is simulated based on the trip information from the data set. We also perform experiments using synthetic data set where demand is generated based on different arrival rate. We use multiple combinations of features such as: (a) Demand-to-Agent-Ratio (DAR): the average number of demand per time step per agent; (b) trip pattern: the average length of trips can be uniform for all the zones or there can be few zones which get longer trips (non-uniform trip pattern); and (c) demand arrival rate: arrival rate of demand can either be static w.r.t. the time or it can vary with time (dynamic arrival rate). At every time step (decision and evaluation point in the simulator), the simulator assigns a trip to the agents based on the number of agents present at the zone and the customer demand. Also, demand expires if it is not assigned within few time steps.

As agents try to maximize their long term revenue, we also provide mean reward of agents (with respect to the time) as the learning progresses and show that VMQ learn policy which yield higher mean values. The mean reward plots are for the running average of mean payoff of all the agents for every 1000 time steps.

Figure \ref{real-world} show results for simulation based on the real-world data set. Plot in Figure \ref{agent-mean} show that agents earn $\approx$5-10\% more value than NFSP and MFQ. Boxplots in Figure \ref{agent-var} exhibit that the variance in the values of individual agents is minimum for VMQ. As agents are playing their best response policy and variance in values is low as compared to other algorithms, we can say that VQM learn policy which is closer to the equilibrium policy.

Figure \ref{synthetic} show results for synthetic data set where we include results for various combination of features. Figures \ref{dar4mean} and \ref{dar4var} plot mean reward and variance in values of agents for a setup with dynamic arrival rate, non-uniform trip pattern with DAR=0.4. The mean reward for VMQ is $\approx$8-10\% higher that NFSP and MFQ. Figures \ref{dar5mean} and \ref{dar5var} show results for a setup with dynamic arrival rate, uniform trip pattern and DAR=0.5. VMQ outperforms NFSP and MFQ by $\approx$5-10\% in terms of average mean payoff of all the individual agents. 
Comparison for an experimental setup with static arrival rate, non-uniform trip pattern and DAR=0.6 is shown in Figures \ref{dar6mean} and \ref{dar6var}. Similar to other setups, mean reward for VMQ is $\approx$5-10\% more than NFSP and MFQ respectively. For all the setups the variance in values of individual agents is  minimum for VMQ. Hence VMQ provides better approximate equilibrium policies. 
\section{Conclusion}
We propose a Stochastic Non-atomic Congestion Games (SNCG) model to represent anonymity in interactions and infinitesimal contribution of individual agents for aggregation systems. We show that the values of all the agents present in a local state are equal at equilibrium in SNCG. Based on this property we propose VMQ which is a \textit{centralized learning decentralized execution} algorithm to learn approximate equilibrium policies. Experimental results on multiple domain depict that VMQ learn better equilibrium policies than the other state-of-the-art algorithms.
\bibliographystyle{named}
\bibliography{vmq}

\end{document}